\documentclass[pdflatex,sn-mathphys-num]{sn-jnl}

\usepackage{graphicx}%
\usepackage{multirow}%
\usepackage{amsmath,amssymb,amsfonts}%
\usepackage{amsthm}%
\usepackage{mathrsfs}%
\usepackage[title]{appendix}%
\usepackage{xcolor}%
\usepackage{textcomp}%
\usepackage{manyfoot}%
\usepackage{booktabs}%
\usepackage{algorithm}%
\usepackage{algorithmicx}%
\usepackage{algpseudocode}%
\usepackage{listings}%
\usepackage{bm}
\usepackage{latexsym}
\usepackage{hyperref}
\usepackage{afterpage}

\theoremstyle{thmstyleone}%
\theoremstyle{thmstyletwo}%

\theoremstyle{thmstylethree}%

\begin{document}

\title{Unifying Information-Theoretic and Pair-Counting Clustering Similarity}

\author*[1]{\fnm{Alexander} J. \sur{Gates}}\email{agates@virginia.edu}

\affil*[1]{\orgdiv{School of Data Science}, \orgname{University of Virginia}, \orgaddress{\street{1919 Ivy Rd}, \city{Charlottesville}, \postcode{22903}, \state{VA}, \country{USA}}}

\abstract{
Comparing clusterings is fundamental to clustering and community detection, yet similarity measures often disagree on the same partitions. 
Existing indices are typically divided into two families: pair-counting measures, which compare co-assignment over element pairs, and information-theoretic measures, which aggregate dependence between clusters. 
We develop an analytical framework relating these families through a shared independence baseline and their underlying sampling spaces.
Mutual information admits an expansion in contingency-table residuals whose leading term is a marginally weighted quadratic departure from independence, while Rand-type indices depend on unweighted co-assignment residuals and marginal-size baselines. 
We further extend pair-counting from unordered pairs to $k$-tuples, showing that pair agreement is the order-two case of a broader collision-counting hierarchy connected to R\'enyi entropies. 
Controlled examples show how these weighting and order differences explain divergence among Rand, adjusted Rand, mutual information, normalized mutual information, and higher-order tuple summaries. 
This unification turns metric disagreement into interpretable evidence about the clustering structures each measure emphasizes.}


\maketitle

\section{Introduction}

The comparison of clusterings is fundamental to evaluating and interpreting unsupervised clustering and community detection models, informing model selection, external validation, ensemble integration, and the longitudinal study of structural evolution across datasets and time \cite{Jain1999data, StrehlGhosh2002JMLR, Meila2003COLT, Meila2007JMVA, Vinh2010JMLR, Danon2005JStatMech, Hric2014communitygroundthruth, kolchinsky2015modularity}. 
Yet, in practice, different families of similarity indices often disagree, sometimes dramatically, which obscures interpretation and decision–making. 
The two most widely used families are 
\emph{pair-counting} indices (e.g., Rand, Adjusted Rand, Jaccard, Fowlkes–Mallows, Mirkin, Wallace) that score agreement over element pairs \cite{Rand1971randindex, HubertArabie1985, FowlkesMallows1983, Steinley2004, Romano2016JMLR}, and 
\emph{information-theoretic} indices (e.g., Mutual Information, its normalizations, and Variation of Information) that aggregate evidence from the full contingency table of cluster co-occurrences \cite{Meila2003COLT, Meila2007JMVA, Vinh2010JMLR, RosenbergHirschberg2007, newman2020improved}.
The resulting plurality of metrics, adjustments, and normalizations has created a landscape in which the same pair of clusterings can be deemed ``similar'' or ``dissimilar'' depending on the index of choice \cite{Pfitzner2009clusteringpairs, Vinh2010JMLR, vanDerHoefWarrens2019, gates2019element, Warrens2022JoC, jerdee2025nmi}.

A central reason for these divergences is what the families emphasize.
Pair-counting indices reduce comparison to the $2\times2$ pair table (same/same, diff/diff, etc.) and therefore implicitly weight each element pair equally; as a consequence, large clusters dominate the score while structure involving minority clusters is often attenuated \cite{Steinley2004, He2009learning,  Warrens2022JoC}. 
On the other hand, information-theoretic indices operate on the full clustering contingency table, where each cell’s contribution is modulated by its expected mass under independence. 
Departures are thus weighted relative to their expected frequency, tending to highlight small but systematic overlaps (often minority–minority intersections) \cite{Danon2005JStatMech, vanDerHoefWarrens2019, Warrens2022JoC, jerdee2024mutual}
In other words, pair-counting measures reward broad overlap of large clusters, while information-based measures are more sensitive to sharp alignments in small ones.

Recognizing these divergent sensitivities, researchers have repeatedly sought to place clustering similarity measures within a common theoretical framework. 
Early work focused on axiomatic properties, identifying desirable features such as invariance under relabeling, sensible normalization, and consistent behavior under cluster refinement and merging, while also showing that no single measure can satisfy all such requirements simultaneously \cite{Meila2005axiom}.
Complementing this, information-theoretic approaches connected clustering comparison to generalized entropy measures, with the Variation of Information providing a metric on the space of probability partitions \cite{Meila2007JMVA}. 
Subsequent work on chance correction showed that both information-theoretic and pair-counting measures can be understood within a common statistical framework based on departures from random assignment \cite{Vinh2010JMLR}. 
From a geometric perspective, many commonly used distances become locally equivalent when two partitions differ only slightly, implying that much of the observed disagreement between measures arises from differences in their global weighting of partition structure rather than their local behavior \cite{meila2012local}. 
Related studies have also interpreted pair-counting measures geometrically in the space of contingency tables \cite{Romano2016JMLR}, and connected them to residual decompositions and cluster-level contributions \cite{vanDerHoefWarrens2019, Warrens2022JoC}, emphasizing the substantial common structure shared by these indices.

These unification efforts suggest that many clustering similarity measures differ less in their underlying structure than in the aspects of partition agreement they emphasize. 
The central challenge for unification is therefore not merely to catalog relationships between indices, but to identify the weighting principles that give rise to their differing sensitivities. 
In this view, clustering measures can be understood as alternative projections of the same contingency-table structure, each emphasizing different scales or forms of agreement.
Unification then becomes an effort to map these sensitivities onto explicit weighting choices over the contingency table, so that particular notions of similarity emerge as tunable regimes rather than fundamentally incompatible behaviors. 
Framed this way, long-standing issues such as normalization, chance correction, and sampling variability can be treated within a common framework; the weighting schemes underlying classical indices become transparent; and agreements or disagreements between measures can be predicted from first principles rather than discovered only through examples. 
Much prior reconciliation has been empirical, comparing indices across synthetic examples or simulation benchmarks to evaluate stability, robustness, or interpretability \cite[e.g.,][]{Pfitzner2009clusteringpairs, amigo2009comparison, Romano2016JMLR, vanDerHoefWarrens2019, Warrens2022JoC}. 
An analytical unification, by contrast, turns discrepancies between measures into interpretable consequences of their mathematical construction, helping practitioners choose indices appropriate for tasks such as ensemble consensus, temporal change detection, or evaluation under severe class imbalance.

Building on this earlier work, we develop a unified view of clustering similarity measures from two complementary perspectives. 
First, we show that both pair-counting and information-theoretic indices can be derived from the same underlying dependence structure on the clustering contingency table. 
The difference between them lies primarily in how departures from independence are weighted and aggregated. 
Pair-counting measures correspond to uniformly weighted, low-order approximations that emphasize broad pairwise consistency, while information-theoretic measures apply probability-weighted higher-order contrasts that are more sensitive to concentrated or localized forms of agreement.

Second, we extend the standard pair-counting framework beyond unordered pairs to higher-order $k$-tuples by adopting a recently introduced family of collision-based approximations to mutual information in the setting of clustering comparison. 
In this formulation, conventional pair agreement appears as the order-two member of a broader hierarchy of dependence measures. 
Pairs, triples, and higher-order tuples represent increasingly stringent notions of co-assignment consistency, while the associated collision probabilities connect naturally to R\'enyi entropies and related information-theoretic quantities. 
This perspective complements recent element-centric approaches based on progressively longer relationships between elements \cite{gates2019element}, while retaining the contingency-table formulation that guarantees invariance under relabeling of clusters.

For clarity, we develop the theory explicitly for the Rand index and mutual information, from which many other pair-counting and information-theoretic measures follow by the same argument. 
The resulting framework explains when the two classes of measures agree and when they diverge. 
Clusterings dominated by large, well-aligned groups tend to favor the uniformly weighted behavior of pair-counting indices, whereas settings with strong class imbalance or small but coherent intersections place greater emphasis on the probability-weighted structure captured by information-theoretic measures. 
In this way, disagreements between indices arise not from fundamentally different notions of similarity, but from explicit choices about weighting, sampling units, and approximation order.

Our framework makes three main contributions. 
First, it provides an analytic bridge between pair-counting and information-theoretic measures through a common independence baseline. 
Second, it extends pair-counting to a hierarchy of higher-order tuple collisions connected to R\'enyi entropy. 
Third, through controlled examples, it shows how weighting and tuple order explain systematic differences among RI, ARI, MI, NMI, and related higher-order measures.
Together, these results recast disagreement between clustering similarity measures as interpretable evidence about which structures each measure emphasizes, rather than as an arbitrary consequence of index choice.

\section{Background and Notation}
\label{sec:background-notation}

\subsection{Clusterings}

We first explicitly introduce a clustering of elements. 
Given a set of $N$ distinct elements $V=\{v_1,\ldots,v_N\}$ (e.g., data points or network vertices), a \emph{clustering} is a partition of $V$ into a family $\mathcal{C}=\{C_1,\ldots,C_{K_{\mathcal{C}}}\}$ of $K_{\mathcal{C}}$ nonempty, pairwise-disjoint subsets (the clusters) such that
\begin{enumerate}\itemsep2pt
  \item $\forall\, i\neq j:\; C_i\cap C_j=\varnothing$,
  \item $\bigcup_{k=1}^{K_{\mathcal{C}}} C_k = V$.
\end{enumerate}
Let $c_k=|C_k|$ denote the size of cluster $C_k$, so the cluster-size sequence is $[c_1,\ldots,c_{K_{\mathcal{C}}}]$.

Throughout, we study the similarity of two clusterings over the same $N$
labeled elements: $\mathcal{A}=\{A_1,\ldots,A_{K_{\mathcal{A}}}\}$ and $
\mathcal{B}=\{B_1,\ldots,B_{K_{\mathcal{B}}}\}$,
with cluster sizes $a_i=|A_i|$ and $b_j=|B_j|$. 
The contingency table $\mathcal{T}$ between two clusterings, shown in Table \ref{tbl:cont}, is $K_{\mathcal{A}}\times K_{\mathcal{B}}$ with cell counts $n_{ij} \;=\; |A_i\cap B_j|$.

\begin{table}[t!]
    	\begin{tabular}{c|c c c c|c}
	$\mathcal{A} / \mathcal{B}$ & $B_1$ & $B_2$ & $\hdots$ & $B_{K_{\mathcal{B}}}$ & Sums \\ \hline
	$A_1$ & $n_{11}$ & $n_{12}$ & $\hdots$ & $n_{1K_{\mathcal{B}}}$ & $a_1$ \\
	$A_2$ & $n_{21}$ & $n_{22}$ & $\hdots$ & $n_{2K_{\mathcal{B}}}$ & $a_2$ \\
	$\vdots$ & $\vdots$ & $\vdots$ & $\ddots$ & $\vdots$ &$\vdots$ \\
	$A_{K_{\mathcal{A}}}$ & $n_{K_{\mathcal{A}}1}$ & $n_{K_{\mathcal{A}}2}$ & \ldots & $n_{K_{\mathcal{A}}K_{\mathcal{B}}}$ & $a_{K_{\mathcal{A}}}$ \\ \hline
	Sums & $b_{1}$ & $b_{2}$ & $\hdots$ & $b_{K_{\mathcal{B}}}$ & $\sum_{ij}n_{ij} = N$
	\end{tabular}
	\caption{The contingency table $\mathcal{T}$ for two clusterings $\mathcal{A} = \{A_1, \ldots, A_{K_{\mathcal{A}}}\}$ and $\mathcal{B} = \{B_1,\ldots, B_{K_{\mathcal{B}}}\}$ of $N$ elements, where $n_{ij} = |A_i\cap B_j|$ are the number of elements that are in both cluster $A_i\in\mathcal{A}$ and cluster $B_j\in\mathcal{B}$.}
	\label{tbl:cont}
\end{table}

\subsection{Clustering similarity and single elements}

To place pair-counting and information-theoretic clustering similarity methods on the same footing, it helps to be explicit about the sampling experiment each one summarizes.
For the information–theoretic family the experiment is simple: pick one element at random and record its two cluster labels.
Specifically, let $u\sim\mathrm{Unif}(V)$ be a uniformly random element of the ground set $V$, and let $i$ (resp.\ $j$) be the cluster index of $u$ under clustering $\mathcal{A}$ (resp.\ $\mathcal{B}$). 
The corresponding probabilities are just normalized counts:
\[
p_{ij}=\frac{n_{ij}}{N},\qquad
\sum_{i=1}^{K_\mathcal{A}}\sum_{j=1}^{K_\mathcal{B}} p_{ij}=1.
\]
The marginals are the row and column sums of the joint,
\[
p_{i\cdot}=\sum_{j}p_{ij}=\frac{a_i}{N},\qquad
p_{\cdot j}=\sum_{i}p_{ij}=\frac{b_j}{N},
\]
so $\sum_i p_{i\cdot}=\sum_j p_{\cdot j}=1$.

The mutual information between two clusterings, $I(\mathcal{A};\mathcal{B})$, is then given in terms of this joint probability:
\[
I(\mathcal{A};\mathcal{B})
=\sum_{i,j}p_{ij}\log\frac{p_{ij}}{p_{i\cdot}p_{\cdot j}}.
\]
This can also be written in terms of clustering entropy terms $I(\mathcal{A};\mathcal{B}) = H(\mathcal{A})+H(\mathcal{B})-H(\mathcal{A};\mathcal{B})$, where $H(\mathcal{A})=-\sum_i p_{i\cdot}\log p_{i\cdot}$ and $H(\mathcal{B})=-\sum_j p_{\cdot j}\log p_{\cdot j}$ are the Shannon entropies of the marginal label distributions for $\mathcal{A}$ and $\mathcal{B}$, respectively, and $H(\mathcal{A};\mathcal{B}) = -\sum_{ij} p_{ij}\log p_{ij}$ is the entropy of the joint distribution.
Similarly, the Variation of Information between two clusterings, $VI(\mathcal{A};\mathcal{B})$, is given by:
\[
VI(\mathcal{A},\mathcal{B})=H(\mathcal{A})+H(\mathcal{B})-2I(\mathcal{A};\mathcal{B}).
\]
Normalized variants such as Normalized Mutual Information (NMI) rescale mutual information relative to the entropy of the clusterings, while Adjusted Mutual Information (AMI) additionally corrects for the expected mutual information under a random model of clusterings \cite{Albatineh2006correctionchance, Gates2017impact}.

\subsection{From single elements to unordered pairs}
\label{sec:background-pair-info}

Pair–counting similarity measures take a different route: they average over unordered pairs of distinct elements sampled uniformly without replacement from the $\binom{N}{2}$ distinct pairs.
This change of sampling space matters: the basic events are now co-assignment versus separation of an element pair.

Formally, pair–counting measures are functions of four numbers:
\begin{align}
A&=\sum_i \binom{a_i}{2}\ \ (\text{pairs co-assigned by }\mathcal{A}), \\
B&=\sum_j \binom{b_j}{2}\ \ (\text{pairs co-assigned by }\mathcal{B}), \\
T&=\sum_{i,j}\binom{n_{ij}}{2}\ \ (\text{pairs co-assigned by both}), \\
M&=\binom{N}{2}\ \ (\text{total pairs}).
\end{align}
Inclusion–exclusion gives the number of pairs separated by both partitions as $M-A-B+T$.

One of the most prominent of the pair-counting similarity measures is the Rand Index \cite{Rand1971randindex}, found as the fraction of element pairs on which the two partitions agree, either both the ``same'' or both ``different’’:
\[
RI=\frac{T+(M-A-B+T)}{M}.
\]
The adjusted Rand index subtracts the chance agreement implied by the fixed marginals and rescales by the maximum possible improvement. 
Under the fixed–marginals (hypergeometric) random model, $\mathbf{E}[T]=AB/M$, which leads to
\[
ARI=\frac{T-\tfrac{AB}{M}}{\tfrac{1}{2}(A+B)-\tfrac{AB}{M}}.
\]
In probability terms these formulas are just $RI=\Pr(\text{agree})$ and $ARI=\big(\Pr(\text{agree})-\Pr_0(\text{agree})\big)/\big(1-\Pr_0(\text{agree})\big)$, where $\Pr_0$ denotes the independence baseline determined by the observed cluster sizes.

Two other commonly used indices emphasize the positive class of co-assigned pairs, one has
\[
\text{Jaccard}=\frac{T}{A + B -T},
\qquad
\text{Fowlkes--Mallows}=\sqrt{\frac{T}{A}\cdot
\frac{T}{B}}.
\]
All of these measures are functions of the same unordered–pair sampling space,
differing only in how they weight its four outcomes.

\section{Independent Clusterings}
\label{sec:independence}

In comparing two clusterings it helps to have a neutral point of reference.
The simplest choice is the independence model: the two clusterings carry no information about one another. 
It fixes the observed cluster sizes but otherwise destroys any structure between them. 
This gives us a clean measuring stick against which to measure departures.

Specifically, the independent (maximum-entropy) pair of clusterings is characterized by
\begin{equation}
\label{eq:indep}
p^{ind}_{ij} \;=\; p_{i\cdot}\,p_{\cdot j}.
\end{equation}

Any observed structure must therefore appear as a deviation from
\eqref{eq:indep}:
\begin{equation}
\label{eq:resid}
\delta_{ij} \;=\; p_{ij}-p^{ind}_{ij}.
\end{equation}
Occasionally we may use its normalized form $\varepsilon_{ij}= \delta_{ij}/(p_{i\cdot}p_{\cdot j})$. 
Since all residuals must cancel out, we have $\sum_j \delta_{ij}=0$ and $\sum_i \delta_{ij}=0$.
The residuals are the basic ``signal'' in what follows: all of the information–theoretic quantities we develop are functionals of $\{\delta_{ij}\}$, and the pairwise indices can be rewritten in terms of quadratic combinations of the same objects.

Because pair–counting lives on unordered pairs, the neutral reference should be defined on the same four counts introduced above in Section~\ref{sec:background-pair-info}: $A,B,T,M$.  
Under the fixed–marginals independence model (same cluster-size marginals, no association), the expected number of pairs co-assigned by both partitions is: $\textbf{E}[T]=\frac{AB}{M}$.
The pairwise independence baseline assumes that the events $AA$ and $BB$ are independent for a random pair. 
Hence the $2\times2$ pair table factorizes:
\begin{equation}
\label{eq:pair-indep}
q^{(0)}_{11}=s_{\mathcal{A}}s_{\mathcal{B}},\quad
q^{(0)}_{10}=s_{\mathcal{A}}(1-s_{\mathcal{B}}),\quad
q^{(0)}_{01}=(1-s_{\mathcal{A}})s_{\mathcal{B}},\quad
q^{(0)}_{00}=(1-s_{\mathcal{A}})(1-s_{\mathcal{B}}),
\end{equation}
where $q_{xy}$ is the probability that a random pair is labeled $x$ by $\mathcal{A}$ and $y$ by $\mathcal{B}$ with $x,y\in\{1=\text{same},\ 0=\text{different}\}$.
Departures $\Delta_{xy}=q_{xy}-q^{(0)}_{xy}$ are precisely what chance–corrected pair indices (e.g., ARI) are designed to capture. 
For large $N$, $s_{\mathcal{A}}=\sum_i (a_i/N)^2+O(1/N)$ (and similarly for $s_{\mathcal{B}}$), so the with– and without–replacement conventions coincide asymptotically while remaining exactly aligned with the combinatorics used by pair–counting measures at finite $N$.

\section{Bridging Mutual Information and Pair-Counting via Expansion Around Independence}

Our next goal is to express both the mutual information between clusterings $\mathcal{A}$ and $\mathcal{B}$ and the Rand index in terms of the residual from the maximally uninformative baseline in equation~\eqref{eq:indep}.
As we shall see, expanding these clustering similarity measures around the independence baseline serves three purposes central to clustering similarity:
(i) it gives a local approximation that is easy to compute and interpret; 
(ii) it exposes a quadratic form that directly bridges information-theoretic and pair-counting families; and 
(iii) it yields higher-order correction terms that can be compared with the $k$-tuple perspective developed below.

\subsection{Expanding mutual information}

For convenience, rewrite the residual as
\[
p_{ij}=p_{i\cdot}p_{\cdot j}(1+\varepsilon_{ij}),
\qquad
\varepsilon_{ij}= \frac{\delta_{ij}}{p_{i\cdot}p_{\cdot j}} .
\]
Plugging this into the mutual information between $\mathcal{A}$ and $\mathcal{B}$ gives
\begin{align}
I(\mathcal{A}; \mathcal{B})
&= \sum_{i,j} p_{i\cdot}p_{\cdot j}\,(1+\varepsilon_{ij})
       \log\!\big(1+\varepsilon_{ij}\big). 
\label{eq:I-start}
\end{align}
Using the classical power-series identity, valid for $|x|<1$ \cite{AbramowitzStegun1964},
\[
(1+x)\log(1+x) \;=\; x \;+\; \sum_{r=2}^{\infty}
\frac{(-1)^r}{r(r-1)}\,x^{\,r},
\]
and substituting $x=\varepsilon_{ij}$, the linear term vanishes because
\[
\sum_{i,j}p_{i\cdot}p_{\cdot j}\varepsilon_{ij}
=
\sum_{i,j}\delta_{ij}
=
0.
\]
Thus
\begin{align} 
I(\mathcal{A}; \mathcal{B})
&= \sum_{i,j} p_{i\cdot}p_{\cdot j}
\sum_{r=2}^{\infty} \frac{(-1)^r}{r(r-1)}\,\varepsilon_{ij}^{\,r} \notag\\
&= \sum_{r=2}^{\infty} \frac{(-1)^r}{r(r-1)}
\sum_{i,j} \frac{\delta_{ij}^{\,r}}{(p_{i\cdot}p_{\cdot j})^{\,r-1}}.
\label{eq:KL-series}
\end{align}

Equation~\eqref{eq:KL-series} gives an exact expansion of the mutual information about the independence baseline. 
The expansion is inherently local: its validity is controlled not by the magnitude of the mutual information itself, but by the normalized cellwise residuals ($\varepsilon_{ij}$). 
As a result, two clusterings may exhibit substantial mutual information while still lying outside the regime where a low-order truncation is accurate, particularly when agreement is concentrated in cells with small expected mass. 
Conversely, when the residuals remain uniformly small, the leading terms provide a controlled approximation to the full quantity. 
The quadratic term captures the leading weighted departure from independence, while the cubic term gives the first correction associated with asymmetry in the residual structure.

This locality can be made explicit with a simple error bound. 
Let $\eta=\max_{i,j}|\varepsilon_{ij}|$.
If $\eta<1$, then the scalar Taylor series in equation~\eqref{eq:KL-series} converges absolutely in every cell. 
After retaining terms through order $R$, the remainder in the mutual-information expansion is bounded by
\[
|R_R^{\mathrm{MI}}|
\leq
\sum_{i,j}p_{i\cdot}p_{\cdot j}
\sum_{r=R+1}^{\infty}
\frac{|\varepsilon_{ij}|^r}{r(r-1)}
\leq
\frac{\eta^{R+1}}{R(R+1)(1-\eta)}.
\]
The bound is intentionally conservative, but it makes clear what controls the approximation. 
There is no universal value of $I(\mathcal{A};\mathcal{B})$ at which a quadratic, cubic, or quartic truncation becomes reliable. 
What matters is the size and distribution of the normalized residuals. 
Cells with small expected mass are especially important: even modest absolute deviations $\delta_{ij}$ can give large values of $\varepsilon_{ij}$, and hence poor convergence of the truncated series. 
We therefore use the first few terms mainly as an interpretive expansion. 
They show how mutual information weights departures from independence, and which parts of the contingency table contribute most strongly to the score.

The expansion begins at second order because of the vanishing linear contribution. 
Retaining only the leading piece of~\eqref{eq:KL-series} gives
\begin{equation}
\label{eq:MI-second}
I(\mathcal{A}; \mathcal{B}) \;\approx\; \frac{1}{2}\sum_{i,j}\frac{\delta_{ij}^{\,2}}{p_{i\cdot}p_{\cdot j}}
\;\equiv\; \frac{1}{2}\,\chi^2_{\!\mathrm{ind}}(\mathcal{A}; \mathcal{B}),
\end{equation}
so mutual information is locally proportional to the Pearson $\chi^2$ statistic for testing independence of the two clusterings.

Two features of this quadratic expansion are worth emphasizing. 
First, the residuals $\delta_{ij}=p_{ij}-p_{i\cdot}p_{\cdot j}$ encode how much mass moves off the independence surface. 
Squaring and summing aggregates these departures into a single measure of deviation. 
Second, each residual is scaled by $1/(p_{i\cdot}p_{\cdot j})$, which up-weights deviations in cells that are rare under independence and down-weights deviations in large, common overlaps. 
This explains a familiar empirical observation: information-based scores tend to be more sensitive than pair-counting scores to small but systematic alignments of minority clusters \cite{Vinh2010JMLR, vanDerHoefWarrens2019}.

The same approximation transfers immediately to the variation of information:
\[
VI(\mathcal{A}; \mathcal{B}) \;\approx\; H(\mathcal{A})+H(\mathcal{B}) - \chi^2_{\!\mathrm{ind}}(\mathcal{A}; \mathcal{B}),
\]
so, to second order, smaller $VI$ corresponds to a larger weighted quadratic departure from independence.

Keeping one more term from~\eqref{eq:KL-series} yields
\begin{equation}
\label{eq:MI-cubic}
I(\mathcal{A}; \mathcal{B})
\;\approx\;
\frac{1}{2}\sum_{i,j}\frac{\delta_{ij}^{\,2}}{p_{i\cdot}p_{\cdot j}}
\;-\;
\frac{1}{6}\sum_{i,j}\frac{\delta_{ij}^{\,3}}{(p_{i\cdot}p_{\cdot j})^{2}}.
\end{equation}
The cubic correction acts like a skewness term on the residual field. 
It becomes large when departures from independence are concentrated in a few strongly positive or negative cells. 
Its sign is also informative: a dominant positive cluster--cluster alignment produces a negative cubic correction, while a pattern dominated by many small negative residuals produces the opposite. 
In empirical use, a sizable cubic term flags regimes where the simple quadratic picture is incomplete, for example when agreement is driven by a handful of tight intersections that pair-counting indices barely register. 
Fourth and higher terms can be written analogously, but they are increasingly sensitive to small expected masses and poorly estimated high-order residual structure.

The mutual-information expansion can also be viewed conditionally:
\[
I(\mathcal A;\mathcal B)
=
\sum_j p_{\cdot j}
D_{\mathrm{KL}}(P_{\mathcal A\mid B_j}\|P_{\mathcal A}).
\]
The quadratic term is therefore a local chi-square geometry measuring how far the conditional label distributions inside clusters of $\mathcal B$ depart from the marginal label distribution of $\mathcal A$. 
Here we emphasize the contingency-table residual form because it remains symmetric in the two clusterings and connects directly to pair-counting indices.

In the language of clustering similarity, the second-order term measures residual alignment beyond chance, with an emphasis determined by cluster-size imbalance. 
The third-order term reports whether that alignment is concentrated in a few intersections or spread more diffusely. 
Together, they explain why two clusterings can have similar Rand or ARI values yet different information-theoretic scores: the latter respond more strongly to rare but coherent overlaps and to asymmetric residual structure.

\subsection{Connection to pair-counting indices}

The independence baseline also gives a convenient lens for inspecting pair–counting indices. 
It tells us what fraction of pair agreements we should see from marginals alone, and it makes explicit that the extra signal in the Rand/ARI family is quadratic in the same residuals that drive the leading term of mutual information.

The Rand index has the exact decomposition
\[
RI \;=\; 1 - \frac{A+B}{M} + \frac{2T}{M}
\;=\;
\underbrace{\Big(1 - \frac{A}{M} - \frac{B}{M} + \frac{2AB}{M^{2}}\Big)}_{\text{independence (marginal) baseline}}
\;+\;
\underbrace{\frac{2}{M}\Big(T - \frac{AB}{M}\Big)}_{\text{residual beyond independence}}.
\label{eq:randexpand}
\]
where the second expansion comes from adding and subtracting $\tfrac{2AB}{M^2}$.
Recall that under the fixed–marginals independence model, the expected number of pairs co-assigned by both partitions is: $\textbf{E}[T]=\frac{AB}{M}$ so the baseline term is precisely the chance agreement implied by the observed cluster sizes .
In this case, and the residual beyond independence is the single scalar $T-\frac{AB}{M}$, which is exactly the numerator term from the adjusted Rand index.

To relate this to the information–theoretic residuals $\delta_{ij}=p_{ij}-p_{i\cdot}p_{\cdot j}$, we rewrite the pair counts:
\[
\frac{A}{M}=\sum_i \frac{\binom{a_i}{2}}{\binom{N}{2}}
=\frac{N}{N-1}\sum_i p_{i\cdot}^{2}-\frac{1}{N-1},
\quad
\frac{B}{M}=\sum_i \frac{\binom{b_j}{2}}{\binom{N}{2}}
=\frac{N}{N-1}\sum_i p_{\cdot j}^{2}-\frac{1}{N-1}
\]
\[
\frac{T}{M}=\sum_{i,j}\frac{\binom{n_{ij}}{2}}{\binom{N}{2}}
=\frac{N}{N-1}\sum_{i,j} p_{ij}^{2}-\frac{1}{N-1}.
\]
Then we make a large $N$ assumption, such that $\frac{N}{N-1}\rightarrow 1$ and $\frac{1}{N-1}\rightarrow0$ which gives the following approximation for the Rand index:
\begin{equation}
RI \;\approx\;
\underbrace{1-\sum_i p_{i\cdot}^{2}-\sum_j p_{\cdot j}^{2}
+2\Big(\sum_i p_{i\cdot}^{2}\Big)\Big(\sum_j p_{\cdot j}^{2}\Big)}_{\text{marginal baseline}}
\;+\; \underbrace{4\sum_{i,j} p_{i\cdot}p_{\cdot j}\,\delta_{ij}}_{\text{linear term}}
\;+\; \underbrace{2\sum_{i,j}\delta_{ij}^{2}}_{\text{quadratic term}}.
\label{eq:RI-prob}
\end{equation}
Hence the Rand index expanded around the pair–counting residual $T-\frac{AB}{M}$ (equivalently, the ARI numerator) decomposes into three terms: i) the marginal independent baseline; ii) a linear alignment term$\,\sum p_{i\cdot}p_{\cdot j}\delta_{ij}\,$; and iii) an unweighted quadratic term $\,\sum \delta_{ij}^{2}$.

The bracketed marginal baseline depends only on the marginals $\{p_{i\cdot}\}$ and $\{p_{\cdot j}\}$, and hence only on the cluster-size distributions of $\mathcal{A}$ and $\mathcal{B}$, not on how elements are matched across the two clusterings.
It is exactly the Rand index you would expect under the independence coupling $p^{(0)}_{ij}=p_{i\cdot}p_{\cdot j}$ (given the large-$N$ approximation). 
In other words, it is the chance agreement implied by the sizes of the clusters alone. 
When the question is ``how much agreement is there beyond what the marginals predict?'' this term is a constant and can be set aside; the adjusted Rand index (ARI) does precisely this by subtracting the baseline and rescaling using the same marginals.

The second term is linear in the residual, $4\sum_{ij}p_{i\cdot}p_{\cdot j}\,\delta_{ij}$, and captures the direction in which probability mass is shifted relative to the marginals. 
Because it is the inner product $\langle \delta,\;p_{i\cdot}p_{\cdot j}\rangle$, it is positive when the excess mass $\delta_{ij}$ is concentrated in high-marginal cells (large $p_{i\cdot}$ and $p_{\cdot j}$) and negative when it is pushed into low-marginal cells. 
This has two important consequences.
First, if the marginals are balanced (all $p_{i\cdot},p_{\cdot j}$ of similar size) or if the residual matrix happens to be nearly orthogonal to the rank–one matrix $(p_{i\cdot}p_{\cdot j})$, the linear term is small and both the pair-counting and information-theoretic families reduce to their common quadratic core, albeit weighted for MI and unweighted for RI/ARI.
Second, under strong size imbalance, the linear alignment term can be substantial, which helps explain why pair–counting indices may report higher agreement driven by large clusters even when MI/VI---dominated by the quadratic, inverse–marginal weights---remain modest.

The third term is quadratic in the residual, $2\sum_{ij}\delta_{ij}^{\,2}$, and gives a nonnegative, unweighted measure of the overall departure from independence; it vanishes if and only if $p_{ij}=p_{i\cdot}p_{\cdot j}$ for all $i,j$ and grows with the $L^2$ distance of the contingency table from the independent baseline. 
In contrast to the MI expansion which has a similar quadratic, there is no factor of $1/(p_{i\cdot}p_{\cdot j})$, so each cell’s influence scales directly with $\delta_{ij}^2$. 
In effect, the leverage sits on high–mass cells in the contingency table: when a large intersection departs from the independence baseline it drives the score, so pair–counting indices emphasize agreement among large clusters and underweight coherent alignments confined to minority clusters.

\subsection{Why not keep adding residual terms?}
The expansion around independence gives an exact series for $I(\mathcal{A};\mathcal{B})$, and its first two terms already capture most of what we see in practice: 
a weighted quadratic that mirrors the pairwise core, plus a cubic skewness correction. 
One might be tempted to push further and retain quartic, quintic, and higher-order terms. 
In practice this is rarely a good bargain. 
Each successive term scales like $\sum_{ij}\delta_{ij}^{\,r}/(p_{i\cdot}p_{\cdot j})^{\,r-1}$, so when some expected masses $p_{i\cdot}p_{\cdot j}$ are small---as they often are with imbalanced cluster sizes---the denominators amplify finite-sample noise in the residuals and the variance balloons unless one applies heavy smoothing. 
The series also converges in the normalized residuals $\varepsilon_{ij}=\delta_{ij}/(p_{i\cdot}p_{\cdot j})$, so a few rare but coherent overlaps with large $|\varepsilon_{ij}|$ can slow convergence to the point where a handful of extra terms adds algebraic complexity without commensurate accuracy. 
And from an interpretability standpoint the return diminishes: the quadratic term has a clean clustering meaning (pairwise agreement beyond chance, with principled weights) and the cubic term adds a useful directional correction; beyond that the higher-order contributions are hard to explain and harder to diagnose empirically.

It is worth noting why we do not pursue an analogous higher-order expansion for pair-counting indices such as the Rand index. 
Once written in terms of probabilities, RI decomposes exactly into a marginal baseline, a weighted linear term, and an unweighted quadratic term in the residuals, with only $O(1/N)$ combinatorial corrections if one keeps the exact $\binom{\cdot}{2}$ terms. 
There is no meaningful hierarchy of structural higher-order terms to uncover: any further expansion refines the finite-sample algebra rather than revealing new clustering effects. 
In short, for MI the higher orders are numerically fragile and conceptually opaque; for RI they are unnecessary. 
This is why we stop at the leading terms and, instead of chasing more residual coefficients, change coordinates entirely in the next section---moving to R\'enyi entropies and collision probabilities that summarize agreement among small tuples in a stable, interpretable way.

\subsection{Toy example: same pair agreement, different contingency geometry.}
\label{sec:toy}

\begin{figure}[t]
	\centering
	\includegraphics[width = 0.7\columnwidth]{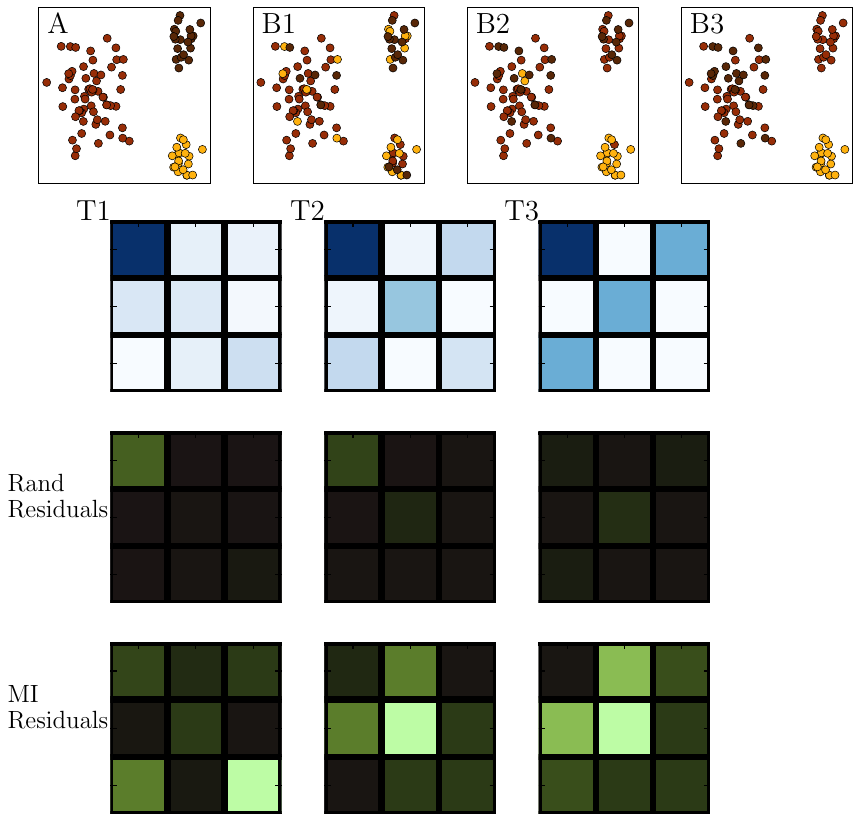}
	\caption{
\textbf{Toy example}: same pair agreement, different contingency-table geometry. 
The first row shows a reference clustering \(\mathcal A\) and three comparison clusterings \(\mathcal B_1,\mathcal B_2,\mathcal B_3\). 
The second row shows the corresponding contingency tables \(T_1,T_2,T_3\). 
All three comparisons have the same marginal cluster sizes and the same aggregate pair-overlap count. 
Consequently, pair-counting measures that depend only on these aggregate pair counts assign identical scores to the three examples, including $RI$, $ARI$, and the order-two collision contrast $I^{(2)}$. 
However, the full contingency-table geometry differs: the same pair agreement is diffuse in \(T_1\), more concentrated in a coherent minority intersection in \(T_2\), and arranged into sharper structured intersections in \(T_3\). 
These differences are detected by information-theoretic and higher-order tuple measures, yielding different \(MI\), \(NMI\), and \(I^{(3)}\) values (see Table \ref{tab:toy-example-scores}). 
The third row shows Rand-style residual contributions, while the fourth row shows mutual-information residual contributions, illustrating how pair-counting compresses distinct overlap geometries that information-theoretic measures retain.
}
	\label{fig:toy}
\end{figure}

To make the loss of contingency-table information explicit, consider four clusterings, $A, B1, B2, B3$ (see Figure \ref{fig:toy}), each over $100$ elements with the same marginal cluster sizes, $(60,20,20)$.
Now consider the following three contingency tables made by comparing clustering $A$ to $B1$, $B2$, and $B3$, respectively:
\[
T_1=
\begin{pmatrix}
49 & 6 & 5 \\
9 & 8 & 3 \\
2 & 6 & 12
\end{pmatrix},
\qquad
T_2=
\begin{pmatrix}
46 & 2 & 12 \\
2 & 18 & 0 \\
12 & 0 & 8
\end{pmatrix},
\qquad
T_3=
\begin{pmatrix}
40 & 0 & 20 \\
0 & 20 & 0 \\
20 & 0 & 0
\end{pmatrix}.
\]
The resulting contingency tables have the same marginals and the same aggregate pair-overlap count, so RI and ARI are identical across the three comparisons: $RI=0.677$ and $ARI=0.342$. 
Nevertheless, the tables encode different overlap geometries. 
In the first comparison, the signal is dominated by the large cluster, with weaker and more diffuse agreement among the smaller clusters. 
In the second, a smaller intersection becomes more coherent. 
In the third, the same amount of pair agreement is arranged into sharper structured intersections. 
Thus the total number of co-assigned pairs is insufficient to describe how the two partitions overlap.

The residual maps make this distinction explicit. 
The Rand-style residuals summarize where the pair-counting signal is located, but the final RI or ARI score collapses these cell-level differences into the same aggregate value. 
The mutual-information residuals retain the location of overlap relative to the factorized baseline \(p_{i\cdot}p_{\cdot j}\). 
As a result, information-theoretic measures distinguish cases in which agreement is concentrated in unexpectedly specific intersections from cases in which the same pair agreement is spread more diffusely. 
This is the central sense in which the full contingency table contains information that pair-counting measures discard: it records not only how many pairs agree, but where those agreements occur relative to the marginal cluster sizes.

\section{Collision Probabilities as a Tuple-Counting Bridge}
\label{sec:renyi-intro}

The residual expansion around independence gives a local view of clustering similarity: it shows how information-theoretic and pair-counting indices aggregate deviations from the same independence baseline. 
There is, however, a complementary route that begins from the pair-counting side. 
Pair-counting measures ask whether two sampled elements receive the same or different cluster labels. 
The natural extension is to ask the same question for larger samples: how often do triples, quadruples, or more generally $k$-tuples of elements receive the same cluster label, or the same pair of labels under two clusterings?
This tuple-sampling view embeds pair-counting in a broader hierarchy of collision statistics and provides a direct bridge to R\'enyi entropies.

\subsection{From Pair Counts to Tuple Counts}

Using the notation introduced above, let $p_{i\cdot}$ and $p_{\cdot j}$ denote the marginal cluster-label distributions and let $p_{ij}$ denote the normalized contingency table for clusterings $\mathcal{A}$ and $\mathcal{B}$. 
For clustering $\mathcal{A}$, the order-$k$ collision probability is $C_k(\mathcal{A}) = \sum_i p_{i\cdot}^{\,k}$, the probability that $k$ independent draws from the element set all receive the same label under $\mathcal{A}$.
Similarly, $C_k(\mathcal{B}) = \sum_j p_{\cdot j}^{\,k}$
For the joint clustering labels, the corresponding collision probability is $C_k(\mathcal{A},\mathcal{B}) = \sum_{i,j}p_{ij}^{\,k}$, the probability that the $k$ sampled elements fall in the same contingency-table cell. 
Equivalently, all $k$ elements receive the same label under $\mathcal{A}$ and the same label under $\mathcal{B}$.

For $k=2$, these quantities recover the population analogues of the pair-counting terms already used by the Rand family:
\begin{equation}
C_2(\mathcal{A})=\sum_i p_{i\cdot}^2,\qquad
C_2(\mathcal{B})=\sum_j p_{\cdot j}^2,\qquad
C_2(\mathcal{A},\mathcal{B})=\sum_{i,j}p_{ij}^2.
\end{equation}
At finite $N$, the without-replacement versions are exactly the familiar binomial ratios,
\begin{equation}
\widehat C_2(\mathcal{A})=\frac{A}{M},\qquad
\widehat C_2(\mathcal{B})=\frac{B}{M},\qquad
\widehat C_2(\mathcal{A},\mathcal{B})=\frac{T}{M}.
\end{equation}
Thus classical pair-counting is the order-two case of a more general collision-counting hierarchy. 
For arbitrary $k$, the finite-population analogues are
\begin{equation}
\widehat C_k(\mathcal{A})
=
\frac{\sum_i\binom{a_i}{k}}{\binom{N}{k}},
\qquad
\widehat C_k(\mathcal{B})
=
\frac{\sum_j\binom{b_j}{k}}{\binom{N}{k}},
\qquad
\widehat C_k(\mathcal{A},\mathcal{B})
=
\frac{\sum_{i,j}\binom{n_{ij}}{k}}{\binom{N}{k}}.
\end{equation}
Pairs ask whether two elements agree; triplets ask whether agreement persists across groups of three; higher orders ask whether larger subsets remain coherent under both clusterings.

\subsection{Connection to R\'enyi Entropy}

Collision probabilities are also the power sums underlying R\'enyi entropy. 
For $\alpha>0$, $\alpha\neq1$, $H_\alpha(\mathcal{A}) = \frac{1}{1-\alpha} \log C_\alpha(\mathcal{A})$, with the Shannon entropy recovered as $\alpha\to1$. 
The corresponding entropy contrast for two clusterings is $K_\alpha(\mathcal{A};\mathcal{B}) = H_\alpha(\mathcal{A}) + H_\alpha(\mathcal{B}) - H_\alpha(\mathcal{A},\mathcal{B})$, or equivalently
\begin{equation}
K_\alpha(\mathcal{A};\mathcal{B})
=
\frac{1}{1-\alpha}
\left[
\log C_\alpha(\mathcal{A})
+
\log C_\alpha(\mathcal{B})
-
\log C_\alpha(\mathcal{A},\mathcal{B})
\right].
\end{equation}
This sign convention gives the Shannon limit
\begin{equation}
\lim_{\alpha\to1}K_\alpha(\mathcal{A};\mathcal{B})
=
I(\mathcal{A};\mathcal{B}).
\end{equation}
We refer to $K_\alpha$ as a R\'enyi entropy contrast rather than a R\'enyi mutual information because several inequivalent R\'enyi generalizations of mutual information exist \cite{vanErvenHarremoes2014}.

For integer orders, this contrast defines a family of tuple-based clustering similarity measures. 
We write $I^{(k)}(\mathcal{A};\mathcal{B}) \equiv K_k(\mathcal{A};\mathcal{B})$ for $k=2,3,\ldots$, so that
\begin{equation}
I^{(k)}(\mathcal{A};\mathcal{B})
=
\frac{1}{1-k}
\left[
\log C_k(\mathcal{A})
+
\log C_k(\mathcal{B})
-
\log C_k(\mathcal{A},\mathcal{B})
\right].
\end{equation}
When measured on a finite clustering, we use the finite-population collision probabilities defined above, yielding
\begin{equation}
\widehat I^{(k)}(\mathcal{A};\mathcal{B})
=
\frac{1}{1-k}
\left[
\log \widehat C_k(\mathcal{A})
+
\log \widehat C_k(\mathcal{B})
-
\log \widehat C_k(\mathcal{A},\mathcal{B})
\right].
\end{equation}
Thus $I^{(k)}$ is an order-$k$ collision contrast: it summarizes agreement using the probability that $k$ sampled elements collide in the same cluster under each marginal clustering and in the same contingency-table cell jointly. 
The order-two case is closest to classical pair-counting because it is built from the same pair-collision quantities underlying the Rand family. 
Higher orders replace pair collisions with triplet, quartet, and larger tuple collisions, producing increasingly stringent summaries of co-assignment consistency.

This contrast shares the same independence baseline as the residual expansion. 
When \(p_{ij}=p_{i\cdot}p_{\cdot j}\), the collision probabilities factor as $C_\alpha(\mathcal{A},\mathcal{B}) = C_\alpha(\mathcal{A})C_\alpha(\mathcal{B})$, and therefore $K_\alpha(\mathcal{A};\mathcal{B})=0$.
Consequently, $I^{(k)}(\mathcal{A};\mathcal{B})=0$ for every integer $k\geq2$ under the factorized contingency table. 
Thus the tuple-counting hierarchy and the residual expansion are calibrated to the same null structure: independence of the two cluster-label distributions.

This shared baseline should not be read as a claim that $I^{(k)}$ is a $k$-term truncation of the mutual-information residual expansion in Eq.~\eqref{eq:KL-series}, or as a generally optimal estimator of Shannon mutual information. 
Although both constructions are centered at the same independence point, they expand different objects. 
Writing \(p_{ij}=q_{ij}(1+\varepsilon_{ij})\), with \(q_{ij}=p_{i\cdot}p_{\cdot j}\), the mutual-information expansion is a Taylor series of
\[
\sum_{ij}q_{ij}(1+\varepsilon_{ij})\log(1+\varepsilon_{ij}),
\]
whereas the order-\(k\) collision contrast depends on
\[
C_k(\mathcal A,\mathcal B)
=
\sum_{ij}q_{ij}^{\,k}(1+\varepsilon_{ij})^k.
\]
Thus \(I^{(k)}\) has its own expansion around independence, but with \(q_{ij}^{\,k}\)-weighted collision moments rather than the \(q_{ij}\)-weighted coefficients of the KL expansion. 
It should therefore be interpreted as an exactly defined order-\(k\) collision summary of the contingency table, not as a measure that shares the first \(k\) terms of the mutual-information series.

In this paper, $I^{(k)}$ is used as a tuple-based clustering similarity family. 
It summarizes agreement through \(k\)-element co-assignment events and provides a finite-resolution bridge between pair-counting and entropy-based comparison. 
Its relationship to Shannon mutual information can be studied through the R\'enyi entropy path, but the accuracy of extrapolating from integer orders \(\alpha=2,3,\ldots\) to the Shannon point \(\alpha=1\) depends on the shape of that path and is not guaranteed by tuple order alone. 
A fuller treatment of low-order collision statistics as finite-resolution approximations to Shannon entropy and mutual information is developed in \cite{gates2026collisioninfo}. 
Here, the important point is that the measures \(I^{(2)},I^{(3)},I^{(4)},\ldots\) remain count-based and interpretable as tuple agreement while also connecting pair-counting to the entropy of the full contingency-table distribution.

\subsection{What Higher-Order Counts Add}

The point of the tuple-counting view is not that a fixed tuple order replaces mutual information. 
Rather, it shows that pair-counting is only the first nontrivial order in a broader sampling hierarchy. 
At order two, the comparison is pairwise and recovers the sampling space of the Rand index, adjusted Rand index, Jaccard index, Fowlkes--Mallows index, and related measures. 
At order three, the comparison asks whether triples remain jointly coherent across the two partitions. 
At higher orders, the statistic increasingly emphasizes whether larger subsets of elements occupy the same cluster intersection.

This hierarchy has a direct information-theoretic interpretation. 
The order-$k$ collision probability is the $k$th power sum of the relevant cluster-label distribution. 
For a single clustering, it is a moment of the marginal cluster-size distribution; for two clusterings, it is a moment of the joint contingency distribution. 
Thus, moving from pairs to triples to larger tuples adds progressively higher-order moment information about the marginal and joint probability distributions induced by the two partitions. 
These moments are exactly the quantities that define integer-order R\'enyi entropies, and therefore provide finite-order observations of the R\'enyi entropy path.

From this perspective, higher-order tuple counts improve the information-theoretic approximation by revealing more of the shape of the probability distribution. 
A pair count captures only quadratic concentration: it tells us how often two elements collide in the same cluster or same contingency-table cell. 
Triplet and quartet counts add cubic and quartic concentration information, distinguishing distributions that have the same pair-collision probability but different higher-order structure. 
Because Shannon entropy and mutual information depend on the full distribution rather than on a single moment, adding higher-order collision statistics can improve finite-resolution approximations to these quantities by capturing additional curvature in the R\'enyi entropy or entropy-contrast path.

In clustering terms, this means that higher-order counts capture whether apparent pairwise agreement is supported by coherent groups. 
Two clusterings may agree on many pairs because of large clusters, yet differ in whether triples or larger subsets remain consistently co-assigned. 
Higher-order collisions therefore help distinguish diffuse pairwise agreement from concentrated group-level alignment. 
This is precisely the structural regime in which pair-counting and information-theoretic measures can diverge: pair-counting records broad co-assignment, while entropy-based measures respond to the distribution of mass across the full contingency table.

The collision hierarchy therefore provides a sampling-based bridge between pair-counting and entropy-based comparison. 
Pair-counting summarizes the lowest-order co-assignment structure. 
Information-theoretic measures operate on the full contingency table. 
Higher-order collision probabilities sit between these views: they remain count-based and interpretable as tuple agreement, but they also add higher-order moment information about the marginal and joint cluster-label distributions. 
In this sense, they provide finite-resolution approximations to the information-theoretic picture while retaining the combinatorial intuition of pair-counting.

\subsection{Toy example: same pair agreement, different contingency geometry.}

The same example from Section \ref{sec:toy} also clarifies the role of higher-order tuple counts. 
Because the three comparisons have identical marginals and identical order-two collision structure, all pair-counting scores are identical, and the order-two collision contrast \(I^{(2)}\) is also identical. 
However, the order-three contrast \(I^{(3)}\) separates the examples, showing that the same pair agreement can be supported by different triplet structure. 
Diffuse pair agreement does not necessarily extend to coherent triples, whereas sharper intersections reinforce the same pair agreement at higher tuple orders.

\begin{table}[t]
\centering
\caption{
Similarity scores for the toy examples in Figure~\ref{fig:toy}.
All three comparisons have identical marginals and identical pair-overlap count, so RI, ARI, and \(I^{(2)}\) are unchanged. 
Mutual information, NMI, and \(I^{(3)}\) distinguish the different contingency-table geometries.
}
\label{tab:toy-example-scores}
\begin{tabular}{lcccccc}
\toprule
Comparison & RI & ARI & MI & NMI & $I^{(2)}$ & $I^{(3)}$ \\
\hline
A vs. B1 & 0.677 & 0.342 & 0.207 & 0.218 & 0.442 & 0.350 \\
A vs. B2 & 0.677 & 0.342 & 0.367 & 0.386 & 0.442 & 0.422 \\
A vs. B3 & 0.677 & 0.342 & 0.568 & 0.598 & 0.442 & 0.539 \\
\hline
\end{tabular}
\end{table}

Table~\ref{tab:toy-example-scores} makes the distinction explicit. 
The order-two collision contrast $I^{(2)}$ behaves like the pair-counting measures because it depends only on pair-collision information. 
The order-three contrast $I^{(3)}$ changes across the examples because triplet collisions detect whether the same pair agreement is diffuse or concentrated into coherent group-level intersections. 
This provides a concrete sense in which higher-order tuple counts recover information that is lost when the comparison is compressed to pair counts alone.

\section{Linking collision probabilities and element–centric similarity}

So far, collision probabilities have placed pair-counting within a broader tuple-sampling hierarchy: pairs measure co-assignment, while triples and larger tuples measure increasingly stringent forms of group-level coherence. 
We now connect this discrete sampling view to the diffusion-based element-centric similarity framework \cite{gates2019element}, which generalizes co-assignment through random walks on element-affinity graphs.

Element–centric similarity was devised to handle overlapping and hierarchical structure, but the core ideas are clearest in the special case of strict partitions (no overlaps). 
In this setting the cluster–affiliation graph breaks into disjoint connected components---one per cluster---and its element–affinity matrix $W$ is block–diagonal.
A random walk on this graph started from a uniformly random element remains within its initial block; the probability that it stays in that block for  $t$ steps is therefore exactly the $(t+1)$-tuple collision probability (all $(t+1)$ draws with replacement landing in the same cluster).
Therefore, we can use a similar trick as above to approximate the personalized PageRank vector
\[
\pi_u \;=\; (1-\alpha)\sum_{t=0}^{\infty}\alpha^t e_u^\top P^t,
\]
using only paths up to order $k$, which reduces exactly to the $k$–tuple collision hierarchy in the partitioned case, where $P$ is the normalized element–affinity matrix $P=D^{-1}W$.
Truncating at length $k$ yields a $k$–path approximation,
\[
\pi_{u}^{(k)} \;=\; (1-\alpha)\sum_{t=0}^{k}\alpha^t e_u^\top P^t,
\]
which captures all paths of length up to $k$ with geometric weights.  
The quantity $\sum_{v}\pi_{u,(k)}(v)$ restricted to $u$’s own cluster then recovers the $k$–tuple collision probability with replacement.

This $k$-tuple expansions reveals how element–centric similarity redirects the focus relative to the two expansions developed above: where the Rand family and mutual information privilege abundance and statistical rarity, respectively, the element–centric view imposes a geometric decay with sample size, sharpening sensitivity to short, transitive structure in the element graph.  
Specifically, in the $k$-path view, personalized PageRank mixes $k$-tuple events with a geometric kernel $((1-\alpha)\alpha^{k-1})$: pairs dominate, triplets are downweighted by $\alpha$, quartets by $\alpha^{2}$, and so on, yielding a tunable locality scale that privileges low-order clustering structures.  
By contrast, the mutual–information expansion around independence emphasizes statistical rarity: its quadratic core weights deviations as $\delta_{ij}^{2}/(p_{i\cdot}p_{\cdot j})$, amplifying coherent but low–mass overlaps irrespective of graph radius.  
The Rand family sits at the opposite extreme: after subtracting the marginal baseline, its leading signal is an unweighted quadratic in $\delta_{ij}$ plus a linear alignment term, making it most responsive to agreement concentrated in large intersections.  
The three schemes therefore select different regimes—local transitivity (geometric (k)-mixing), rare systematic alignment (MI/VI), and bulk agreement (RI/ARI)—and the appropriate choice depends on which mode of structural coherence one wishes to detect.

The random-walk formulation can therefore be viewed as a generalization of tuple-based comparison beyond strict partitions. 
For hard, non-overlapping clusterings, tuple collisions measure whether small groups of elements remain jointly co-assigned. 
For overlapping or hierarchical clusterings, element-affinity graphs allow two elements to be related through multi-hop membership chains even when they do not share a single cluster directly. 
The $k$-path expansion provides the corresponding notion of scale: short paths capture local coherence, analogous to low-order tuple collisions, while longer paths reflect broader connectivity in the element-affinity graph. 
Thus, collision probabilities and element-centric similarity share a common intuition---structural agreement can be probed by increasing the sampling radius---but they implement it in different ways: one through explicit tuple co-assignment events, the other through geometrically weighted random-walk paths.

\section{Illustrative Examples}

The theoretical development so far identifies two complementary sources of disagreement among clustering similarity measures. 
First, pair-counting and information-theoretic indices weight contingency-table residuals differently: pair-counting measures compress agreement into aggregate pair totals, whereas information-theoretic measures retain where overlap occurs relative to the independence baseline. 
Second, the tuple-counting hierarchy shows that pair agreement is only the order-two case of a broader sequence of collision summaries, where higher orders test whether pairwise agreement is supported by coherent triples and larger groups. 
In this section, we use controlled examples to illustrate these mechanisms. 
The examples separate the effects of residual weighting, cluster-size imbalance, and tuple order, showing when RI, ARI, MI, NMI, and the collision contrasts \(I^{(k)}\) agree, diverge, or reveal different aspects of the same clustering comparison.

We design a small set of controlled experiments that isolate the regimes highlighted by our theory: (i) abundance vs.\ rarity (Rand vs.\ MI weighting), (ii) locality (geometric $k$–mixing), and (iii) chance calibration. 
In all examples we compare RI, ARI, MI, and the $k$–tuple approximations $I^{(2)}$, $I^{(3)}$, $I^{(4)}$. 
To place all curves on a common $0–1$ scale we report a normalized score by dividing by the average of the clustering self-similarity.
Specifically, for similarity measure $S(\mathcal{A},\mathcal{B})$ we divide by $\left(S(\mathcal{A},\mathcal{A}) + S(\mathcal{B},\mathcal{B})\right)/2$ so that the measure always equals $1$ at perfect agreement.

Our first stylized clustering example places $N=1000$ elements into a balanced clustering $\mathcal{A}$ with two clusters of size $500$.
From $\mathcal{A}$ we generate a second clustering $\mathcal{B}$ by exchanging the membership of a fraction of the elements ($\epsilon\in[0,0.5]$).
For each $\epsilon$ we compute RI, ARI and normalized versions of $MI$, $I^{(2)}$, $I^{(3)}$, and $I^{(4)}$.
Each point represents the mean over 100 independent random trials, and the shaded areas indicate two standard errors of the mean.

\begin{figure}[t]
	\centering
	\includegraphics[width = 0.7\columnwidth]{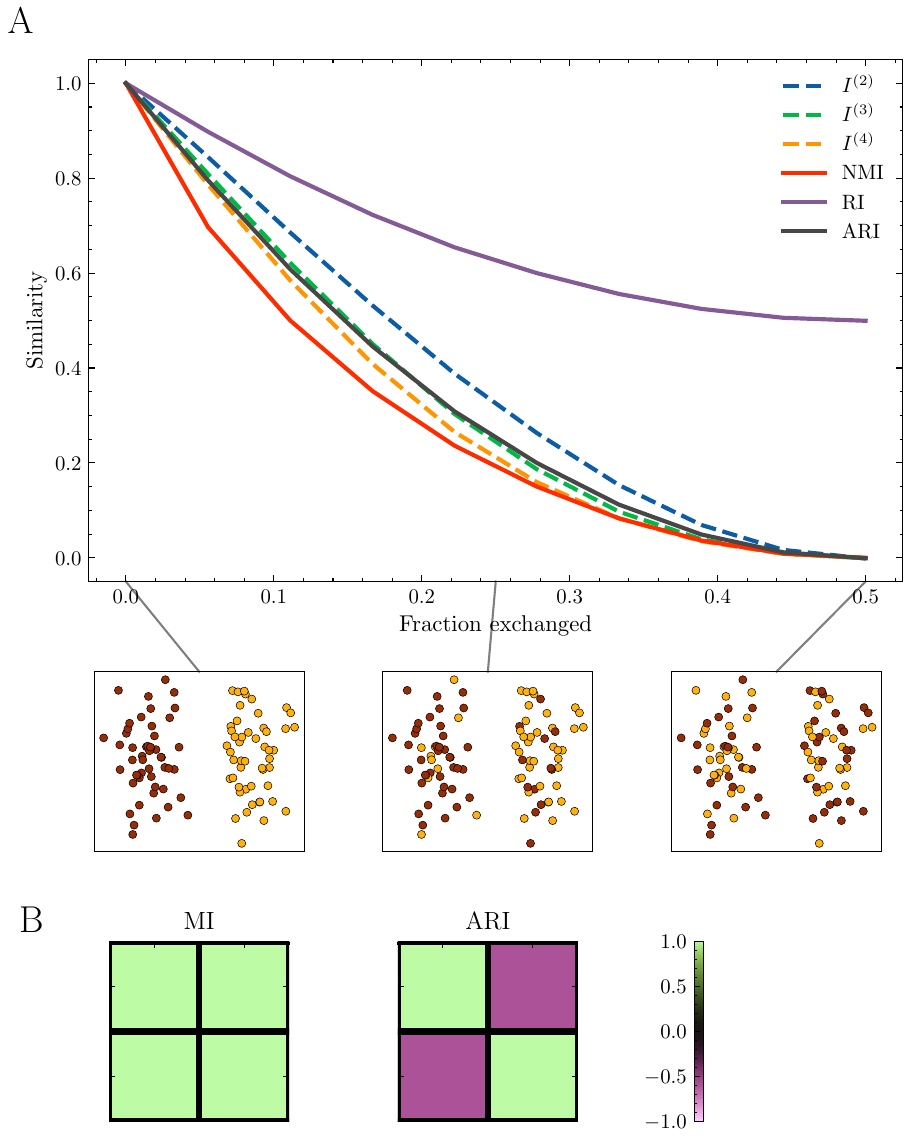}
	\caption{\textbf{Balanced clustering similarity.}
    $N=1,000$ elements are grouped into a balanced clustering $\mathcal{A}$ with two clusters of size $500$, while $\mathcal{B}$ is made by exchanging the membership of a fraction of the elements, $\epsilon\in[0,0.5]$, from $\mathcal{A}$.  
    \textbf{A}) For each $\epsilon$ we compute the Rand Index (RI), Adjusted Rand Index (ARI), Normalized Mutual Information (NMI), and normalized variants of the $I^{(2)}$, $I^{(3)}$, and $I^{(4)}$ Rényi contrasts approximations to MI.
    Curves represent the average over 100 independent trials, while shaded area reflects two standard errors of the mean.
    \textbf{B}) Residual matrices (normalized to highlight relative magnitudes) for the MI and ARI between $\mathcal{A}$ and $\mathcal{B}$ with exchange level $\epsilon=0.25$.}
	\label{fig:balanced}
\end{figure}

In this first, balanced, experiment, the measures separate cleanly by what they weight (Figure \ref{fig:balanced}A). 
The Rand index (RI) falls the slowest because it counts all pairs uniformly: many pairs remain untouched even as labels are perturbed, and by $\varepsilon=\tfrac12$ the assignment is effectively random, yielding the well-known RI baseline of about $0.5$ (half the pairs agree by chance). 
Normalized mutual information (NMI) drops the fastest: with two balanced labels the mapping is a binary symmetric channel and shrinks sharply from the top and reaches $0$ at $\varepsilon=\tfrac12$, reflecting the complete loss of predictability of one label from the other. 
The $k$-tuple approximations sit neatly between these extremes, with a monotone ordering $I^{(2)} > I^{(3)} > I^{(4)})$: increasing $k$ discounts short, accidental pair matches and rewards higher-order consistency, so the curves descend more quickly toward the MI trajectory as $k$ grows.
Finally, ARI closely tracks $I^{(3)}$ in this balanced, symmetric-noise setting rather than $I^{(2)}$.  
Although both are ``pair–based,'' ARI’s chance–corrected signal is an \emph{unweighted} quadratic in broken–pair residuals (without replacement, no logarithm), whereas $I^{(2)}$ is a Rényi–contrast built from collision probabilities passed through a log, yielding a concave mapping that compresses near the top and drops faster in the midrange. 
By contrast, the $k=3$ combination $I^{(3)}$ introduces a triplet term whose contribution, in this regime, behaves effectively like an additional linear correction to pairwise agreement—capturing concentrated, coherent overlaps in a way that mirrors ARI’s linear component.

The residual analysis gives a more detailed view into how each measure weights and aggregates cell-level deviations in the contingency table. 
Whereas global similarity scores summarize overall agreement, the residual patterns reveal the underlying balance of positive and negative contributions that drive those scores. 
As shown in Figure \ref{fig:balanced}B, for this balanced example, the MI receives nearly uniform positive contributions from all cells, reflecting its symmetric treatment of departures from independence. 
In contrast, the ARI shows positive residuals along the diagonal---corresponding to correctly matched clusters---and negative residuals off the diagonal, where elements are split or merged across clusters.
This decomposition illustrates that MI captures overall dependence, while ARI quantifies net pairwise consistency by offsetting agreement against disagreement.

\begin{figure}[t]
	\centering
	\includegraphics[width = \columnwidth]{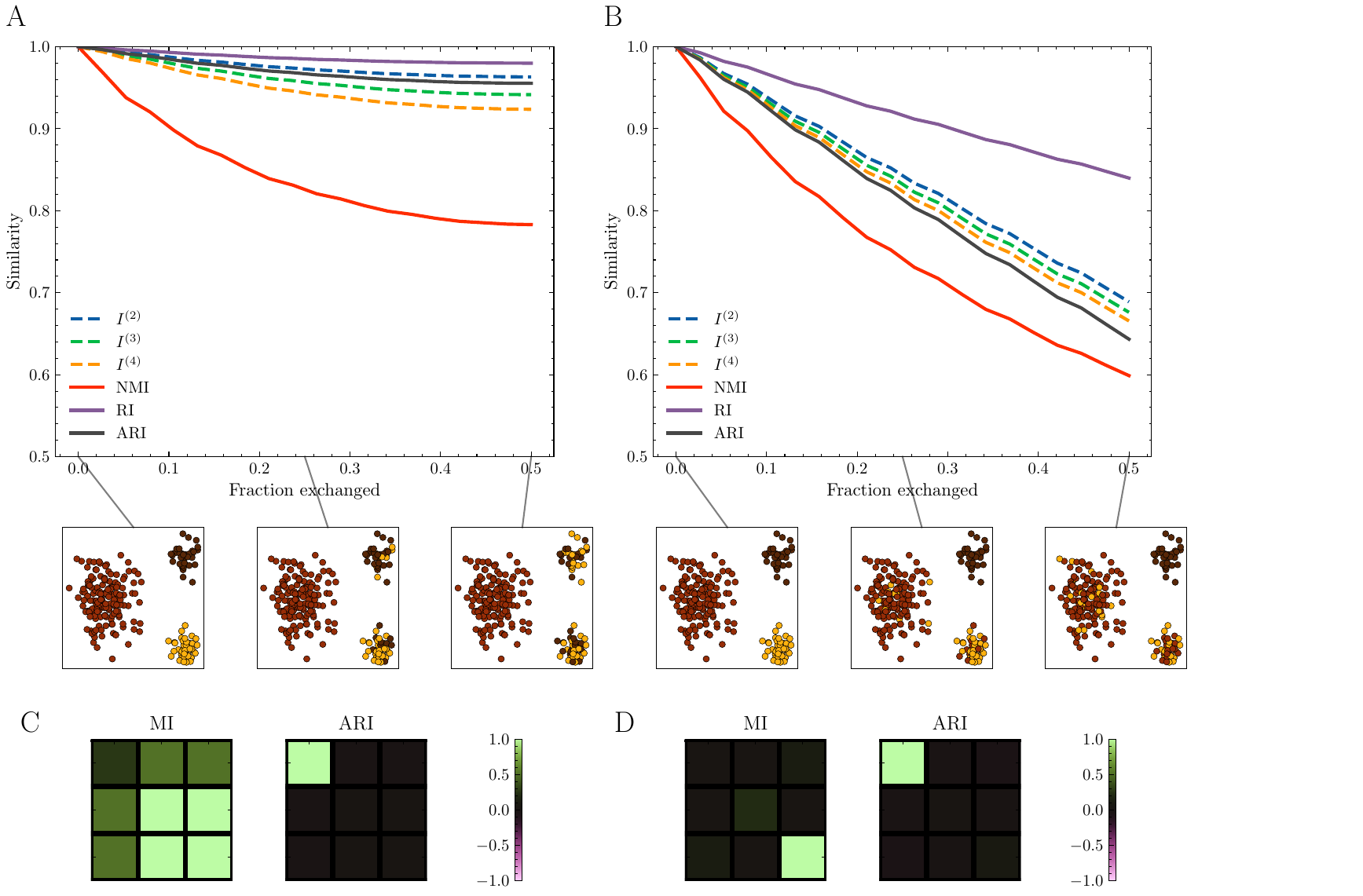}
	\caption{\textbf{Unbalanced clustering similarity.}
    $N=1,000$ elements are grouped into an unbalanced clustering $\mathcal{A}$ with one big cluster of size $800$ and two small clusters of $100$ elements each, while clustering $\mathcal{B}$ is made by exchanging the membership of a fraction of the elements, $\epsilon\in[0,0.5]$, from $\mathcal{A}$ between the (\textbf{A}) ``small-small'' clusters or (\textbf{B}) ``big-small'' clusters.  
    For each $\epsilon$ we compute the Rand Index (RI), Adjusted Rand Index (ARI), Normalized Mutual Information (NMI), and normalized variants of the $I^{(2)}$, $I^{(3)}$, and $I^{(4)}$ Rényi contrasts approximations to MI.
    Curves represent the average over 100 independent trials, while shaded area reflects two standard errors of the mean. 
    \textbf{C}-\textbf{D}) Residual matrices (normalized to highlight relative magnitudes)for the MI and ARI between $\mathcal{A}$ and $\mathcal{B}$ with exchange level $\epsilon=0.5$ for the \textbf{C} small-small and \textbf{D} big-small exchange examples.}
	\label{fig:unbalanced}
\end{figure}

Our second stylized example places $N=1,000$ elements into an unbalanced clustering $\mathcal{A}$ with one large cluster containing 80\% of the elements ($800$), and two small clusters with 10\% each ($100$).
From $\mathcal{A}$ we again generate a second clustering $\mathcal{B}$ by exchanging the membership of a fraction of the elements ($\epsilon\in[0,0.5]$), but this time differentiating between exchanges between the two small clusters (Figure \ref{fig:unbalanced}A) and a small cluster and the large cluster (Figure \ref{fig:unbalanced}B).

When cluster sizes are highly uneven, which disagreements are introduced matters as much as how many; in both cases---the ``small–small'' and ``big–small''' exchanges---we flip the same number of elements, but their impacts differ because the Rényi contrasts and Mutual Information weight deviations by the joint probability mass of the cells they disturb.
In the small–small case (Figure \ref{fig:unbalanced}A), all changes are confined to low–frequency intersections ($p_{i\cdot}p_{\cdot j}$ are tiny), so mutual information drops quickly since its quadratic core scales like $\delta_{ij}^{\,2}/(p_{i\cdot}p_{\cdot j})$.
By contrast, ARI (and $I^{(k)}$) weight by frequency, not inverse marginals, so they register a modest change when only minority–minority cells are perturbed. 
This observation is supported in the residual analysis (Figure \ref{fig:unbalanced}C), which shows how ARI is dominated by the similarity of the big-cluster and receives a barely visible signal of the disagreement between small clusters, while MI has a much more prominent residual signal from the random exchanges between the small clusters.
As $\epsilon\rightarrow0.5$, all curves begin to change curvature, reflecting the symmetry of the setup: since the two small clusters are of equal size and their labels are exchangeable, half the elements swapped corresponds to the point where the two partitions are as far from the original clustering as possible.

In the big–small exchange (Figure \ref{fig:unbalanced}B), the same number of moved elements now disrupts one large, high-mass intersection. 
Because those cells dominate the contingency table, all measures fall much more sharply.
Here the pair–counting perspective dominates: ARI’s leading signal is essentially an unweighted quadratic in broken pairs, which in our runs aligns best with the higher–order collision approximation. 
Again, the residual analysis for ARI reflects how the index is dominated by changes to the big cluster (Figure \ref{fig:unbalanced}D).
On the other hand, the MI residuals reflect inverse-frequency weighting, amplifying deviations involving smaller clusters. 
They highlight the strong alignment of the third cluster, a partial mismatch within the large cluster, and complete disagreement for the smallest cluster.
Unlike the symmetric small–small case, there is no curvature reversal because the exchange is asymmetric: the large cluster cannot be relabeled to restore equivalence, and so similarity continues to decline monotonically.

Overall, the contrast between these two regimes underscores how ARI responds to the mass distribution of disagreements, not merely their count. 
When disruptions are confined to rare, low-mass cells, ARI behaves like the pair-based $I^{(2)}$; when they involve dominant clusters, ARI’s unweighted residual structure aligns with the higher-order $I^{(k)}$, revealing how the same underlying collision framework naturally bridges the two behaviors.

\section{Discussion}

The results presented here clarify the relationship between pair-counting and information-theoretic clustering similarity measures. 
The divide between these families is not eliminated, but it becomes more interpretable. 
Our framework shows that both families can be understood relative to the same independence baseline, differing primarily in how they weight and aggregate departures from that baseline.
Specifically, the residual expansion shows that mutual information, variation of information, and related quantities are locally governed by a marginally weighted quadratic form in the contingency-table residuals. 
Rand-type measures, by contrast, depend on unweighted co-assignment residuals together with marginal-size baselines. 
This difference in weighting explains why information-theoretic measures are more responsive to rare but systematic intersections, while pair-counting measures emphasize broad agreement among large clusters.

Collision probabilities provide a complementary sampling interpretation of the same divide. 
Classical pair-counting is the order-two case of a broader tuple-counting hierarchy: pairs measure co-assignment, while triples, quartets, and larger tuples measure increasingly stringent forms of group-level coherence. 
The corresponding collision probabilities are power sums of the marginal and joint cluster-label distributions, and therefore connect naturally to R\'enyi entropies. 
In this paper, the resulting order-$k$ contrasts \(I^{(k)}\) are used as clustering similarity measures: count-based summaries that remain close to pair-counting in their sampling interpretation while inheriting an entropy-based connection to the full contingency table. 
A full treatment of these collision contrasts as finite-resolution approximations to Shannon entropy and mutual information is developed separately \cite{gates2026collisioninfo}; here, their role is to show how pair-counting fits inside a broader tuple-sampling hierarchy.

The synthetic examples highlight the practical consequences of these two views. 
When clusters are balanced and noise is symmetric, the measures tend to move together because the weighting differences are muted and the same residual structure drives all scores. 
Under strong imbalance, however, the measures diverge. 
Information-theoretic scores respond more strongly to coherent changes in small clusters because their residual weights scale relative to expected contingency-table mass. 
Pair-counting scores remain more stable when the large clusters are preserved, because many element pairs still agree. 
Higher-order tuple summaries sit between these regimes: they remain based on concrete co-assignment events, but they test whether agreement persists beyond pairs into triples and larger groups.

This framework suggests a more precise way to choose among clustering similarity measures. 
If the goal is to summarize coarse agreement among large, relatively homogeneous clusters, pair-counting measures such as RI and ARI are often appropriate. 
If the goal is to detect small but systematic alignments, minority subtypes, or fine-grained overlaps, information-theoretic measures such as MI, NMI, AMI, or VI may be more informative. 
If the goal is to ask whether pairwise agreement is supported by coherent groups, then higher-order tuple contrasts provide an intermediate family. 
These choices should not be treated as interchangeable normalizations of a single notion of similarity. 
They encode different assumptions about which structures matter.

The present analysis focuses on hard partitions of a fixed set of elements. 
In that setting, the contingency table captures the complete relationship between the two clusterings. 
Many empirical clustering problems, however, involve overlapping communities, hierarchical structure, or probabilistic memberships. 
For such cases, element-centric similarity provides a path-based extension \cite{gates2019element}. 
The connection developed here suggests why: tuple collisions probe local group coherence in strict partitions, while random walks on element-affinity graphs probe structural coherence over paths. 
Short paths play a role analogous to low-order tuple events, whereas longer paths incorporate broader connectivity in the affinity graph. 
This offers a route for extending the same sampling logic beyond disjoint partitions.

Several limitations remain. 
First, the residual expansion is most transparent near the independence baseline, where low-order terms dominate. 
When residuals are large or highly concentrated in cells with small expected mass, higher-order terms can become important and the local approximation is less stable. 
Second, high-order tuple counts can be sparse in finite samples, especially for many small clusters or fragmented contingency tables. 
Thus, although higher-order collisions add structural information, they also increase data requirements. 
Third, the R\'enyi contrasts used here should be interpreted as clustering similarity summaries, not as a replacement for the full theory of Shannon mutual information. 
Fixed tuple order defines a finite-resolution population target; increasing the order changes what is being measured.

Several future directions follow. 
One is to develop finite-sample corrections and uncertainty estimates for higher-order tuple contrasts, especially under fixed-marginal permutation nulls. 
A second is to connect the residual and tuple-counting views more directly to statistical testing: the same contingency-table residuals that explain metric disagreement also define natural test statistics for departures from independence. 
A third is to extend the framework to overlapping, hierarchical, and soft clusterings by replacing hard co-assignment events with weighted or probabilistic co-assignment. 
Finally, the collision hierarchy suggests multiscale clustering similarity measures in which tuple order controls the resolution of comparison.

Beyond clustering, the same ideas connect to broader themes in network science and information theory. 
The independence baseline parallels null models used in community detection, and the residuals \( \delta_{ij} \) resemble assortativity terms that quantify deviations from random mixing. 
The collision hierarchy is also analogous to motif or higher-order interaction counts: it asks whether structure persists when one moves from pairwise relations to larger subsets. 
These analogies suggest that the framework may be useful for comparing network partitions, role assignments, embeddings, or other discrete structural summaries.

Overall, the framework developed here does not identify a universally best clustering similarity measure. 
Instead, it explains why different measures disagree. 
Pair-counting and information-theoretic indices share a common baseline, but they weight residual structure differently and operate on different sampling spaces. 
Higher-order collision counts expose an intermediate hierarchy between pairwise agreement and full contingency-table information. 
This turns metric disagreement from a nuisance into evidence about the scale and location of agreement: whether similarity is carried by large clusters, rare but coherent intersections, or increasingly stringent group-level structure.

\section*{Code Availability}
Implementations of all discussed measures and examples are provided through CluSim \cite{Gates2019clusim}; \href{https://github.com/Hoosier-Clusters/clusim}{https://github.com/Hoosier-Clusters/clusim} with a corresponding notebook in the examples folder: UnifyingInfoPair\_ClusteringSimilarity.ipynb.


\section*{Acknowledgements}
The author would like to thank great conversations with students in his research group, the Connected Data Hub.
The author was supported in part by the National Security Data \& Policy Institute, Contracting Activity \#2024-24070100001.

\bibliography{clusteringrefs}

\end{document}